\documentclass[11pt,twocolumn]{article}

\usepackage[margin=1in]{geometry}   % 邊距
\usepackage{times}                  % 字型
\usepackage{setspace}              
\usepackage{graphicx}              % 插圖
\usepackage{amsmath, amssymb}      % 數學
\usepackage{hyperref}              % 超連結
\usepackage{titlesec}              % 控制標題間距
\usepackage[authoryear,round]{natbib}
\usepackage{subcaption}

\titlespacing{\section}{0pt}{*2}{*1}
\titlespacing{\subsection}{0pt}{*1.5}{*0.5}
\usepackage{placeins} 
\usepackage{tabularx}

\title{Human Protein Atlas Image Classification Research}

\usepackage{tabularx}  % 加在 preamble，排表格用

\author{
\centering
\begin{tabular}{c c}
Sylvey Lin & Zhi-Yi Cao \\
\texttt{yuhsinl2@illinois.edu} & \texttt{zihyic2@illinois.edu}
\end{tabular}
}

\date{}

\begin{document}

\begin{titlepage}
    \centering
    \vspace*{2cm}
    
    {\LARGE\bfseries Leveraging Diffusion Models for Synthetic Data Augmentation in Protein Subcellular Localization Classification \par}
    \vspace{1cm}
    
    {\Large Sylvey Lin \quad yuhsinl2@illinois.edu  \par}
    {\Large Zhi-Yi Cao \quad zihyic2@illinois.edu \par}
    \vspace{1cm}

    \begin{minipage}{0.9\textwidth}
    \textbf{Abstract} \par
    \vspace{0.5em}
    We investigate whether synthetic images generated by diffusion models can enhance multi-label classification of protein subcellular localization. Specifically, we implement a simplified class-conditional denoising diffusion probabilistic model (DDPM) to produce label-consistent samples and explore their integration with real data via two hybrid training strategies: Mix Loss and Mix Representation. While these approaches yield promising validation performance, our proposed MixModel exhibits poor generalization to unseen test data, underscoring the challenges of leveraging synthetic data effectively. In contrast, baseline classifiers built on ResNet backbones with conventional loss functions demonstrate greater stability and test-time performance. Our findings highlight the importance of realistic data generation and robust supervision when incorporating generative augmentation into biomedical image classification.
    \end{minipage}

    \vspace{0.5cm}

    \includegraphics[width=0.85\linewidth]{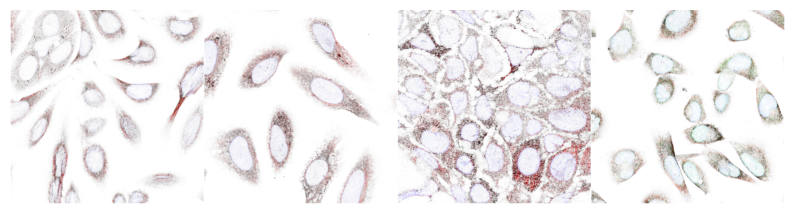} 
    \includegraphics[width=0.85\linewidth]{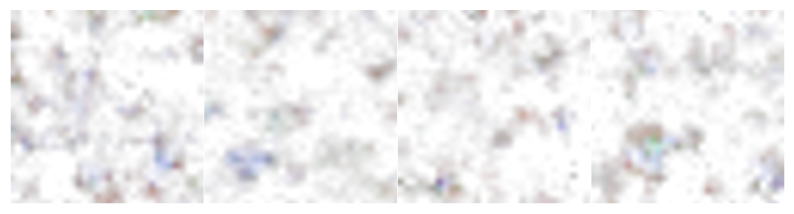} \\
    \textit{Figure: Comparison between original dataset images (top) and diffusion-generated samples (bottom).}
    
    \vspace{0.5cm}
    \begin{center}
        \small The complete source code and diffusion-generated data can be accessed at: \url{https://shorturl.at/bRiuC}
    \end{center}
\end{titlepage}

\section{Introduction}

Proteins are “the doers” in the human cell, executing a wide range of functions that are essential to life. Understanding where proteins localize within cells is a critical step toward deciphering biological processes, identifying disease mechanisms, and enabling breakthroughs in drug discovery and precision medicine. While previous efforts have primarily focused on single-pattern classification within limited cell types, real-world biological systems often exhibit mixed localization patterns across diverse cell populations. Accurately identifying these patterns at scale remains a significant challenge.

To address this, the Human Protein Atlas launched a Kaggle competition aimed at developing models capable of classifying protein subcellular localization from high-throughput fluorescence microscopy images. Each image contains four color channels representing cellular components such as the nucleus, microtubules, and endoplasmic reticulum, in addition to the protein of interest. The competition emphasizes multi-label classification, where proteins may simultaneously localize to multiple subcellular compartments. Evaluation is based on the macro F1 score, which treats all classes equally and amplifies the impact of rare categories—making the task especially sensitive to class imbalance.

The ultimate goal of this challenge is to integrate top-performing models into a smart-microscopy system that automates the identification of protein localization patterns. However, the complexity of protein structures, variable staining quality, and noisy imaging conditions demand robust modeling strategies and often require substantial labeled data. To mitigate this, prior solutions have leveraged extensive data augmentation, external datasets (e.g., Human Protein Atlas v18), and sophisticated techniques such as metric learning with antibody-level identity supervision. For example, the winning solution by the bestfitting team employed DenseNet121 with post-processing, external label mapping, and ArcFace-based embedding learning to achieve superior test-time performance.

Motivated by these successes, our work explores a more constrained yet potentially scalable approach: can we improve classification performance using only the official competition dataset—without access to external labels or curated metadata? Specifically, we propose a two-stage pipeline: (1) use diffusion models to generate class-conditional synthetic images, and (2) train a downstream classifier under a semi-supervised learning framework inspired by MixMatch. We investigate two hybrid training strategies—Mix Loss and Mix Representation—to integrate real and generated data, and evaluate their impact on model generalization. A detailed discussion of related work follows in the next section.

\section{Related Works}
Diffusion models have recently gained traction in data augmentation, offering a powerful means of generating diverse and semantically meaningful training examples. \citet{trabucco2024effective} demonstrate that semantically-guided image transformations using pretrained diffusion models can significantly improve few-shot classification performance across diverse domains. \citet{diffuseMix2024} further propose DIFFUSEMIX, which blends real and synthetic images with structural patterns (e.g., fractals) to preserve label semantics and enhance generalization.

In parallel, traditional image-mixing techniques such as MixUp \citep{mixUp} have shown success by interpolating pairs of input images and their labels to improve model robustness. Inspired by both lines of work, our approach adopts a more direct pairing strategy that mixes real and diffusion-generated samples without relying on handcrafted visual effects or additional augmentation layers.

\section{Dataset}
The Human Protein Atlas (HPA) Image Classification dataset provides high-resolution microscopy images for classifying subcellular protein localization patterns. It was created using standardized confocal microscopy techniques and is widely used for studying protein functions and cellular structures.

The dataset contains 31,072 samples, each represented by four grayscale images corresponding to different fluorescence channels: nucleus (blue), microtubules (red), endoplasmic reticulum (yellow), and the protein of interest (green). The green channel is the primary signal for classification, while the other channels provide structural context. The images are available in 512x512 resolution for computational efficiency and high-resolution formats for more detailed analysis.

Each sample is labeled with one or more of 28 protein localization categories, making this a multi-label classification task. The dataset also spans 27 different cell types, each with distinct morphological features that influence the visual appearance of protein expression. A major challenge is class imbalance, requiring methods such as weighted loss functions or data augmentation to improve model performance.

This dataset serves as a benchmark for deep learning applications in biomedical imaging, particularly in classifying complex protein distributions. It supports the development of automated methods for analyzing cellular structures, with potential applications in computational biology and medical research.

\section{Baseline Methods}
To establish a reliable performance reference, we evaluated several baseline models using only the official competition dataset. Our comparisons spanned various backbone architectures and loss functions, to assess their robustness in the presence of class imbalance and fluorescence image noise.

\subsection{Backbone Architectures}
We experimented with the following backbone models:

\begin{itemize}
        \item \textit{ResNet} variants (e.g., ResNet18, ResNet34), known for their efficiency and stability in image classification tasks.
        \item \textit{Swin Transformer}, a hierarchical vision transformer that processes images in shifted windows, enabling both local and global feature extraction with improved efficiency.
        \item \textit{Vision Transformer (ViT)}, which applies pure transformer blocks to image patches, effectively capturing long-range spatial dependencies through self-attention mechanisms.
\end{itemize}

These architectures were selected to compare convolutional networks and transformer-based models in multi-label classification.

\subsection{Loss Functions}
To address the challenges of class imbalance and overlapping protein patterns, we evaluated the following loss functions:

\begin{itemize}
    \item \texttt{BCEWithLogitsLoss}, a standard choice for multi-label classification.
    \item \texttt{ArcFace}, an additive angular margin loss originally proposed for face recognition (Deng et al., 2019). We adopt this approach with an ArcMarginProduct layer, inspired by the bestfitting team's use of metric learning to enhance class separability.
    \item \texttt{FocalLoss}, designed to address class imbalance by focusing training on hard-to-classify examples.
\end{itemize}

\subsection{Observations}
Some configurations achieved promising results on the local validation set but failed to generalize to the Kaggle public leaderboard. Others struggled even during validation, particularly those using more complex backbones under limited training epochs. These outcomes underscore the difficulty of achieving stable and generalizable performance using only the official dataset, and further motivate the investigation of advanced modeling strategies, such as generative augmentation or semi-supervised learning, to overcome limitations in data diversity and generalization capacity.

\begin{figure*}
    \centering
    \includegraphics[width=1\linewidth]{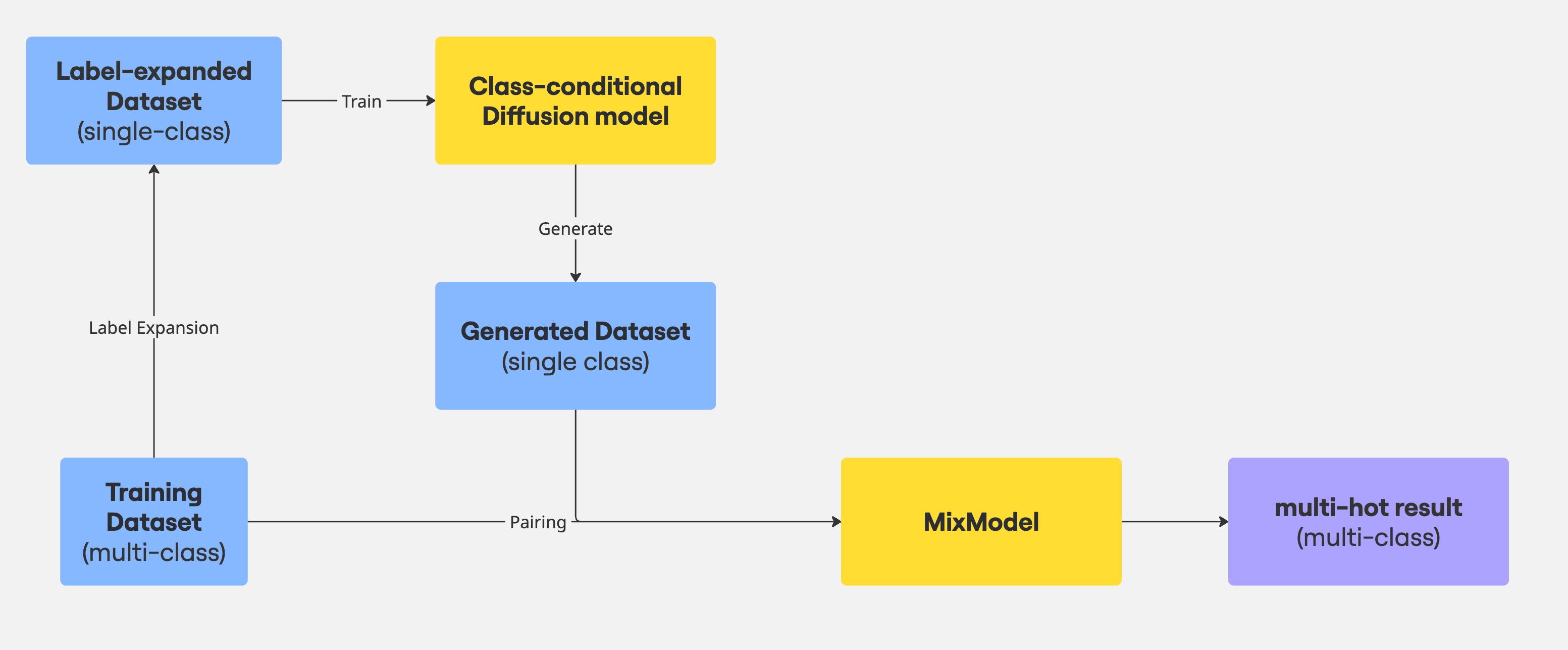}
    \caption{Model Structure Overview}
    \label{fig:model_overiew}
\end{figure*}

\section{Proposed Method}
To address the generalization limitations observed in our baseline experiments, we propose a custom architecture that leverages synthetic data generated by diffusion models to improve training diversity. While prior top-performing solutions relied heavily on external datasets such as the Human Protein Atlas v18, we investigate whether comparable benefits can be achieved through generative augmentation alone, without the use of additional labeled resources. Specifically, we employ a simplified class-conditional DDPM to synthesize label-consistent samples, which are integrated into classifier training via two hybrid strategies: Mix Loss and Mix Representation. An overview of the proposed architecture is provided in Figure~\ref{fig:model_overiew}.

\subsection{Diffusion Model}
\begin{figure*}
    \centering
    \includegraphics[width=0.6\linewidth]{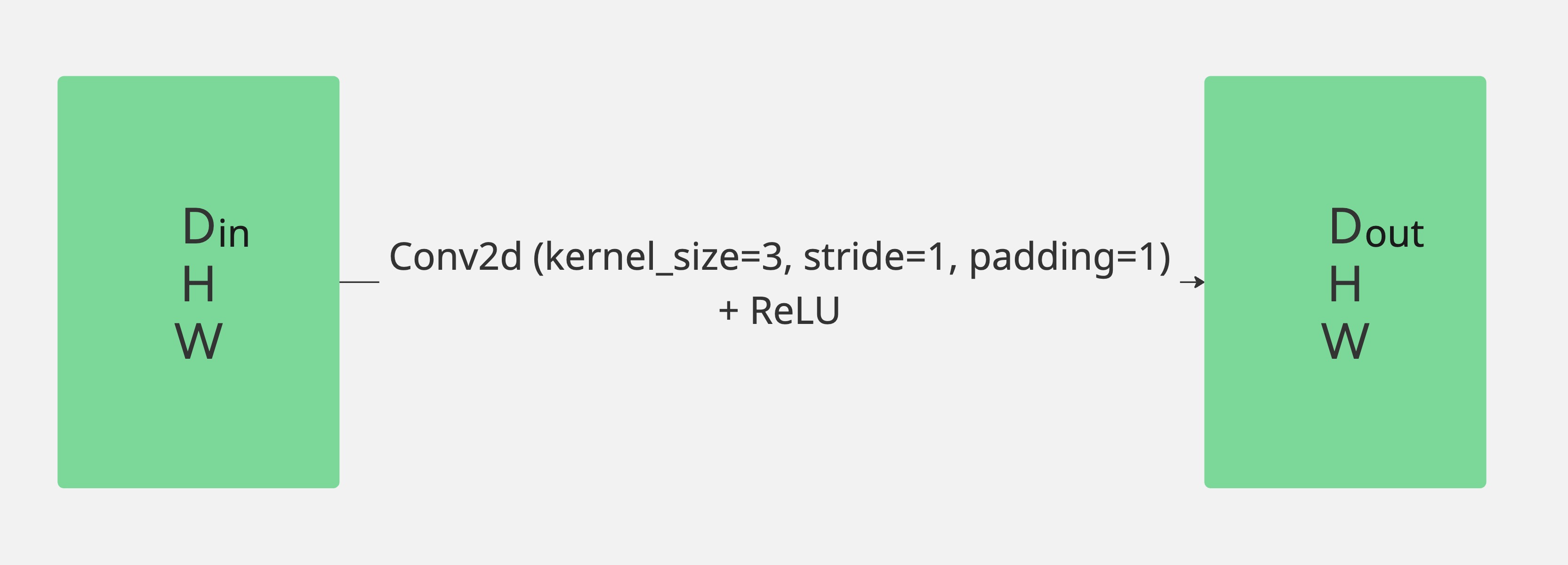}
    \caption{Simplified Conv$D_{\text{out}}$ structure}
    \label{fig:convDout}
    \vspace{2em}
    \includegraphics[width=1\linewidth]{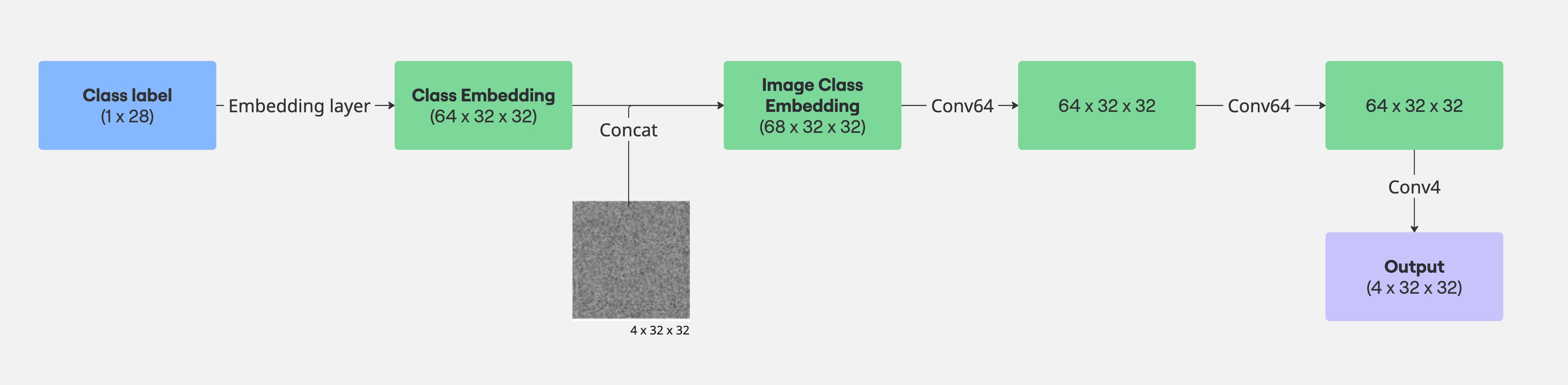}
    \caption{Simplified class-conditional UNet}
    \label{fig:simplified_ddpm}
\end{figure*}

The original class-conditional DDPM \citep{ddpm} was computationally expensive and failed to generate class-distinct outputs under our resource constraints. We therefore design a simplified variant that preserves core denoising capabilities while reducing architectural complexity. Compared to the original UNet—which employs a full encoder–decoder structure, skip connections, multi-resolution convolutions, and extensive conditioning—our model eliminates up/downsampling paths and uses a shallow three-layer convolutional network.

Class conditioning is applied once at the input level by concatenating a class embedding to the image tensor, while timestep conditioning is omitted. We further simplify the design by using standard ReLU activations without batch normalization or attention mechanisms. This lightweight formulation facilitates efficient evaluation of class-conditional generation under limited computational budgets, while retaining the fundamental behavior of a denoising network (see Figures~\ref{fig:convDout} and~\ref{fig:simplified_ddpm}).

To accommodate the model's reduced capacity, input images are downsampled from $512 \times 512$ to $32 \times 32$ during diffusion training. The generated outputs are then upsampled via bicubic interpolation to match the original resolution expected by the downstream classifier.

\subsection{MixModel}
To incorporate both real and synthetic data into training, we adopt a MixUp-inspired framework \citep{mixUp}. For each training instance, a real image and a diffusion-generated image are randomly paired, and both their inputs and labels are linearly combined using coefficients sampled from a Beta distribution:

\begin{align}
\lambda &\sim \text{Beta}(\beta, \beta) \\
x' &= \lambda x_1 + (1 - \lambda) x_2 \\
p' &= \lambda p_1 + (1 - \lambda) p_2
\end{align}

Here, $x_1$ and $x_2$ represent a real and a synthetic image, respectively, and $p_1$, $p_2$ are their label distributions. The mixing coefficient $\lambda$ controls the interpolation ratio; smaller values of $\beta$ favor more polarized combinations. We extend this formulation through two complementary strategies: feature-level mixing and supervision-level mixing.

\subsubsection{Mix Representation}
In this variant, interpolation is performed in feature space rather than input space. Using a shared ResNet backbone (ResNet18 or ResNet34, modified for 4-channel input), we extract intermediate feature representations from both real and synthetic samples. The final classification layer is removed to retain embeddings. A mixing coefficient $\lambda \sim \text{Beta}(\beta, \beta)$ is sampled for each pair:

\begin{align}
f_{\text{mix}} &= \lambda f_r + (1 - \lambda) f_g \\
y_{\text{mix}} &= \lambda y_r + (1 - \lambda) y_g
\end{align}

Where $f_r$, $f_g$ are feature vectors from the real and synthetic images, and $y_r$, $y_g$ their respective labels. The mixed feature vector $f_{\text{mix}}$ is passed to a linear classification head, and training is performed using binary cross-entropy loss on the interpolated target $y_{\text{mix}}$.

\subsubsection{Mix Loss}
As an alternative, we combine model predictions and losses directly. Given real and synthetic samples, the network computes logits $z_r$ and $z_g$ from the shared backbone and classification head. Corresponding binary cross-entropy losses $\mathcal{L}_r$ and $\mathcal{L}_g$ are calculated with respect to $y_r$ and $y_g$. A mixing coefficient $\lambda \sim \text{Beta}(\beta, \beta)$ is sampled and used as follows:

\begin{align}
z_{\text{mix}} &= \lambda z_r + (1 - \lambda) z_g \\
\mathcal{L}_{\text{mix}} &= \lambda \cdot \mathcal{L}_r + (1 - \lambda) \cdot \mathcal{L}_g
\end{align}

Final predictions are derived from $z_{\text{mix}}$, and the training objective is computed as the average of $\mathcal{L}_{\text{mix}}$. This method preserves separate supervision signals for real and synthetic branches and complements the feature-level strategy.

\section{Experiment}

We conduct a series of experiments to evaluate the effectiveness of our proposed method and compare it with supervised baselines. This section details our data split, training environment, baseline configurations, and the training protocol for the proposed diffusion-based approach.

\subsection{Validation Strategy}

To assess generalization performance, we partition the official training set into 90\% for training and 10\% for validation. All reported metrics are based on this held-out validation set unless otherwise noted. In addition, we submit predictions to the Kaggle competition platform to obtain scores on both public and private test splits, offering an external benchmark for evaluating generalization.

\subsection{Environment}

All models were trained using Google Colab Pro+, equipped with high-performance GPUs (primarily A100). This platform facilitated rapid prototyping and allowed us to run experiments interactively and efficiently.

\subsection{Baseline Training Configuration}

To establish strong baselines, we evaluated several ResNet architectures, including ResNet18, ResNet34, and ResNet50, using three different loss functions: BCEWithLogitsLoss (binary cross-entropy), Focal Loss, and ArcFace. These combinations allow us to assess robustness under class imbalance and feature separability challenges.

We trained all baseline models with a batch size of 64 using the AdamW optimizer. Two different learning rate schedules were used depending on the number of training epochs:

\begin{itemize}
    \item \textbf{Scheduler A}: Step decay schedule for shorter training runs (e.g., 30 epochs), reducing the learning rate by a factor of 0.1 every 15 epochs:
    \[
    \text{lr} = \text{init\_lr} \times 0.1^{\left\lfloor \frac{\text{epoch}}{15} \right\rfloor}
    \]
    
    \item \textbf{Scheduler B}: Manually defined multi-stage decay for longer runs (e.g., 80 epochs), adapted from the public implementation by the \textit{bestfitting} team:
    \[
    \text{lr} =
    \begin{cases}
    3.0 \times 10^{-4}, & \text{epoch} \leq 25 \\
    1.5 \times 10^{-4}, & 25 < \text{epoch} \leq 30 \\
    7.5 \times 10^{-5}, & 30 < \text{epoch} \leq 35 \\
    3.0 \times 10^{-5}, & 35 < \text{epoch} \leq 40 \\
    1.0 \times 10^{-5}, & \text{epoch} > 40
    \end{cases}
    \]
\end{itemize}

This staged schedule proved effective in controlling overfitting during extended training.

\subsection{Proposed Method Training}

For the diffusion model, we initially followed the original DDPM training setup using AdamW with an initial learning rate of $3 \times 10^{-4}$, a CosineAnnealingLR scheduler, and 40 training epochs. In our final configuration, using the simplified DDPM structure, we trained for 30 epochs with a fixed learning rate of $1 \times 10^{-4}$ and batch size of 64. The model generated 384 images per class, totaling 10,752 diffusion-generated samples.

For the MixModels, we used both ResNet18 and ResNet34 as backbones, adapted to accept 4-channel input. The mixing coefficient $\lambda$ was sampled from a $\text{Beta}(\beta, \beta)$ distribution with $\beta = 0.3$. Each model was trained with a batch size of 64, a fixed learning rate of $1 \times 10^{-4}$, and the AdamW optimizer. To adjust the learning rate based on validation performance, we applied a ReduceLROnPlateau scheduler (decay factor 0.5, patience 2 epochs). Early stopping with a patience of 5 epochs was employed to mitigate overfitting. Training was capped at a maximum of 100 epochs.

\section{Results and Discussion}

\subsection{Baseline models}
Our baseline experiments focus on evaluating two key factors that can influence model performance: the choice of ResNet backbone architecture and the use of different loss functions, including ArcFace and FocalLoss.

\subsubsection{Comparison of ResNet Backbones}
We first analyze how the choice of ResNet backbone architecture impacts model performance. To ensure a fair comparison, we fixed the learning rate at $1 \times 10^{-3}$, used the BCEWithLogits loss function, and trained all models for 30 epochs. The results are shown in Table~\ref{tab:different_backbones}.

Interestingly, under identical training settings and constrained computational resources, the simpler ResNet18 consistently outperformed ResNet34 across all evaluation metrics. This suggests that, in low-epoch scenarios, smaller models may generalize better due to their faster convergence and reduced risk of overfitting. 

Notably, we observed a substantial performance degradation on the Kaggle leaderboard compared to internal validation. The F1-scores obtained from public and private test sets were significantly lower than those on the validation set, indicating that the models struggled to generalize beyond the training distribution.

\begin{table}[h]
    \centering
    \begin{tabularx}{\linewidth}{l|X|X|X}
        \textbf{Model} & \textbf{Validation} & \textbf{Public Score} & \textbf{Private Score} \\
        \hline\hline
        ResNet18 & 0.6383 & 0.45702 & 0.43363 \\
        ResNet34 & 0.6258 & 0.44622 & 0.42232 \\
    \end{tabularx}
    \caption{Validation and test performance comparison between ResNet18 and ResNet34 with different backbones under fixed learning conditions.}
    \label{tab:different_backbones}
\end{table}

\begin{table*}
    \centering
    \begin{tabular}{l|c|c|c|c|c|c}
        \textbf{Model}  & \textbf{lr} & \textbf{loss func} & \textbf{epochs} & \textbf{Validation} & \textbf{Public Score} & \textbf{ Private Score} \\
        \hline\hline
        ResNet34 & 1e-3 & BCEWithLogitsLoss & 30 & 0.6258 & 0.44622 & 0.42232 \\
        ResNet34 & 1e-3 & ArcfaceLoss & 30 & 0.4263 & 0.33799 & 0.33491 \\
        ResNet50 & 1e-4 & ArcfaceLoss & 80 & 0.5140 &  0.40058 & 0.38895 \\
    \end{tabular}
    \caption{Performance of ArcFace loss under different model capacities, compared to BCE baseline.}

    \label{tab:arcfaceLoss}
    \vspace{2em}
    \begin{tabular}{l|c|c|c|c|c|c}
        \textbf{Model} & \textbf{lr} & \textbf{loss func} & \textbf{epochs} & \textbf{Validation} & \textbf{Public Score} & \textbf{ Private Score} \\
        \hline\hline
        ResNet18 & 1e-3 & BCEWithLogitsLoss & 30 & 0.6383 & 0.45702 & 0.43363 \\
        ResNet18 & 1e-3 & FocalLoss & 30 & 0.6327 &  0.39496 & 0.37423 \\
        ViT-B/16 & 1e-4 & FocalLoss & 10& 0.4209 & 0.01800 &0.01373 \\
        Swin-B & 1e-4 & FocalLoss & 10& 0.6489 & 0.18306 &0.17674 \\
    \end{tabular}
    \caption{Evaluation of Focal Loss on ResNet18 and transformer-based architectures (ViT, Swin).}
    \label{tab:focalLoss}
\end{table*}

\subsubsection{Evaluation of ArcFace Loss}

We also investigated the use of ArcFace. Using the same ResNet34 backbone, learning rate, and training epochs as in the baseline configuration, we found that substituting BCEWithLogitsLoss with ArcFace resulted in a significant drop across all evaluation metrics.

To examine whether this degradation stems from architectural limitations, we replaced the backbone with ResNet50 and adjusted the scheduler, learning rate, and training epochs accordingly. While the ResNet50 model with ArcFace achieved better results than its ResNet34 counterpart, it still underperformed compared to the ResNet34 model trained with BCEWithLogitsLoss.

Table~\ref{tab:arcfaceLoss} summarizes the results across these configurations.

\subsubsection{Evaluation of Focal Loss}

In addition to ArcFace loss, we also investigated the performance of models trained with FocalLoss. Since ResNet18 requires fewer computational resources, we used it to compare FocalLoss against the standard BCEWithLogitsLoss under identical training configurations.

On our internal validation set, FocalLoss achieved performance comparable to BCEWithLogitsLoss, suggesting that it may be effective in handling class imbalance during training. However, when evaluated on the Kaggle public and private test sets, FocalLoss exhibited poor generalization, underperforming the baseline on external test sets.

We further evaluated FocalLoss on two transformer-based models, ViT-B/16 and Swin-B. Both models were trained for 10 epochs due to resource constraints, and no learning rate scheduler was applied. While Swin-B achieved promising results on the validation set, its performance dropped significantly on the Kaggle test sets, suggesting potential overfitting or limited generalization capability. ViT-B/16, on the other hand, performed poorly across all metrics, indicating that the model may require substantially more training or that it is less suited for this task under the given experimental conditions.

Table~\ref{tab:focalLoss} summarizes the results of all models trained with FocalLoss.

\subsection{Proposed Model}
In this section, we present the evaluation results of our proposed pipeline, which combines diffusion-based synthetic data generation with mixed supervision for classification. We first analyze the behavior of the diffusion model, followed by experiments using our MixModel architecture.

\subsubsection{Diffusion Model}
\begin{figure}[h]
    \centering
    \begin{subfigure}[b]{0.45\linewidth}
        \centering
        \includegraphics[width=\linewidth]{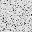}
        \caption{Class 0}
        \label{fig:fig1}
    \end{subfigure}
    \hfill
    \begin{subfigure}[b]{0.45\linewidth}
        \centering
        \includegraphics[width=\linewidth]{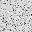}
        \caption{Class 1}
        \label{fig:fig2}
    \end{subfigure}
    \caption{Visualization of the red channel from images generated for different classes by the original class-conditional DDPM model.}
    \label{fig:class0_1ori}
\end{figure}

We first evaluated the original class-conditional DDPM model. When used for image generation, the model failed to produce class-distinct outputs. As shown in Figure~\ref{fig:class0_1ori}, the generated images lack meaningful visual patterns and show minimal resemblance to the original training data. Moreover, under identical noise conditions, the model produced nearly identical outputs across different class labels, suggesting that it failed to properly utilize class-conditioning during generation.

\begin{figure}[h]
    \centering
    \begin{subfigure}[b]{0.45\linewidth}
        \centering
        \includegraphics[width=\linewidth]{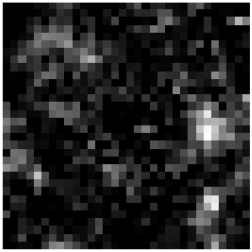}
        \caption{Class 0}
        \label{fig:fig1}
    \end{subfigure}
    \hfill
    \begin{subfigure}[b]{0.45\linewidth}
        \centering
        \includegraphics[width=\linewidth]{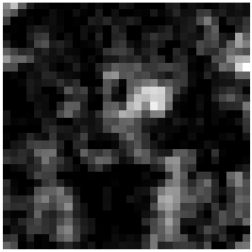}
        \caption{Class 1}
        \label{fig:fig2}
    \end{subfigure}
    \caption{Visualization of the red channel from images generated for different classes by the simplified DDPM model.}
   
    \label{fig:class0_1simpl}
\end{figure}

We then turned to the simplified DDPM model. As shown in Figure~\ref{fig:class0_1simpl}, the generated images exhibit blurry yet semantically meaningful patterns. Moreover, under identical noise conditions, the model is able to produce visibly different outputs for different class labels, suggesting that it successfully leverages class-conditioning during the generation process.

\begin{figure}
    \centering
    \includegraphics[width=1\linewidth]{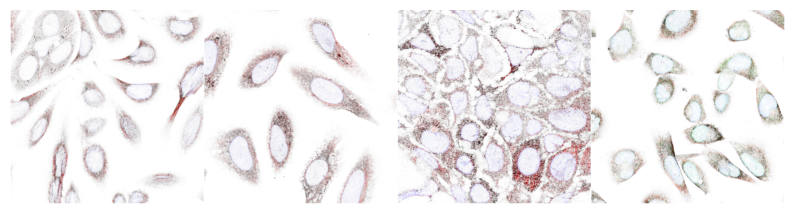}
    \includegraphics[width=1\linewidth]{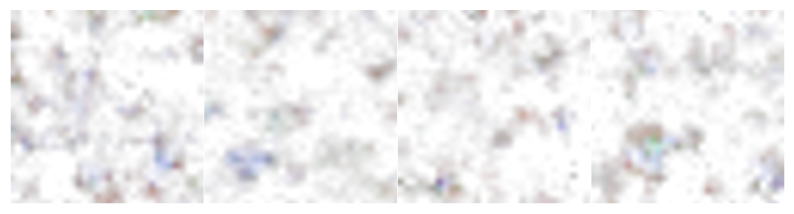}
    \caption{Class-conditional generation results (red channel) showing minimal distinction across labels.}
    \label{fig:Ori_gen}
\end{figure}

Since the simplified DDPM model demonstrated better performance in both capturing visual patterns and differentiating between classes, we adopted it for generating images used in downstream classifier training. Figure~\ref{fig:Ori_gen} illustrates a visual comparison between samples from the original dataset and those generated by the simplified diffusion model.

\subsubsection{MixModel}

To explore whether synthetic images generated by diffusion models could improve classification performance when combined with real data, we experimented with two mixing strategies: \textit{Mix Loss} and \textit{Mix Representation}. These strategies were applied to two different ResNet backbones (ResNet18 and ResNet34), resulting in four model configurations. Table~\ref{tab:ComMix} summarizes their performance on validation, public test, and private test sets.

\begin{table*}[h]
    \centering
    \begin{tabularx}{\linewidth}{l|X|X|X|X}
        \textbf{Model} & \textbf{Mix Method} & \textbf{Validation} & \textbf{Public Score} & \textbf{Private Score} \\
        \hline\hline
        ResNet18 & Mix Loss & 0.2451 & 0.01924 & 0.01955 \\
        ResNet34 & Mix Loss & 0.2953 & 0.01924 & 0.01955\\
        ResNet18 & Mix Representation & 0.3250 & 0.01924 & 0.01955 \\
        ResNet34 & Mix Representation & 0.3027 & 0.01924 & 0.01955 \\
    \end{tabularx}
    \caption{F1-scores on validation and Kaggle test sets for MixModels trained with different backbones and mixing strategies.}
    \label{tab:ComMix}
\end{table*}

Among all MixModel configurations, the best validation score was achieved using \textit{Mix Representation} with a ResNet18 backbone (0.3250), followed closely by ResNet34 with the same strategy. This suggests that mixing at the representation level may better preserve semantic structure during training. However, none of the MixModel variants outperformed the supervised baselines, and all configurations—regardless of mixing method or backbone depth—exhibited equally poor performance on the Kaggle public and private test sets, with F1-scores stagnating around 0.019.

This significant discrepancy between validation and test performance points to a serious generalization failure. We attribute this failure to several contributing factors: (1) overfitting to non-natural patterns introduced by the diffusion process; (2) limited visual diversity in generated samples; and (3) distributional mismatch between real and synthetic features, especially under linear interpolation.

\begin{figure}[h]
    \centering
    \includegraphics[width=1\linewidth]{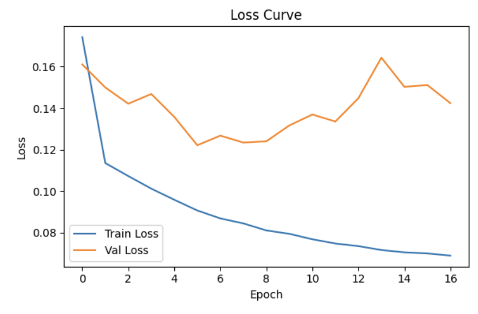}
    \caption{Training and validation loss curve for MixModel with ResNet34 (Mix Loss). The divergence between training and validation loss suggests potential overfitting.}
    \label{fig:mixloss_curve}
\end{figure}

To further investigate this gap, we examined the training and validation loss curves for MixModel with ResNet34 and Mix Loss. As shown in Figure~\ref{fig:mixloss_curve}, the training loss steadily decreases, while the validation loss plateaus and fluctuates early on. This divergence is characteristic of overfitting, where the model continues to optimize on the training data without gaining improvements on unseen validation samples.

\subsection{Summary of Findings}
Overall, while our baseline models achieved moderate validation performance, all configurations—both supervised and semi-supervised—exhibited limited generalization on unseen data. Although synthetic data from diffusion models introduced additional visual diversity, it also posed challenges in distribution alignment and semantic consistency. These findings highlight the difficulty of leveraging generative augmentation in high-resolution, multi-label biomedical image classification tasks without stronger guidance or constraints.

\section{Conclusion}

In this work, we explored the use of diffusion-generated synthetic data to enhance protein subcellular localization classification, combining traditional supervised models with a novel MixModel framework. While baseline models using ResNet backbones and BCEWithLogitsLoss achieved moderate validation performance, their generalization to the Kaggle leaderboard was limited.

Our proposed pipeline incorporated a simplified class-conditional DDPM for image generation, along with two hybrid training strategies—Mix Loss and Mix Representation—to integrate real and synthetic samples. Although MixModels yielded higher validation scores, particularly under feature-level mixing, none of the configurations generalized well to unseen test data, with Kaggle F1-scores remaining consistently low. In all cases, the semi-supervised approaches failed to surpass supervised baselines.

These findings underscore the challenges of leveraging synthetic data in the absence of external resources or extensive tuning. Artifacts introduced by the generative model, insufficient sample diversity, and distributional mismatches likely contributed to the generalization failure.

Nonetheless, this study offers practical insights into the limitations of generative augmentation for high-resolution, multi-label biomedical image classification. Future directions include improving diffusion model fidelity, incorporating stronger conditioning mechanisms, and adopting adaptive training objectives to better align synthetic and real data distributions.

\bibliography{anthology,custom}

\begin{thebibliography}{4}
\providecommand{\natexlab}[1]{#1}
\providecommand{\url}[1]{\texttt{#1}}
\expandafter\ifx\csname urlstyle\endcsname\relax
  \providecommand{\doi}[1]{doi: #1}\else
  \providecommand{\doi}{doi: \begingroup \urlstyle{rm}\Url}\fi

\bibitem[Berthelot et~al.(2019)Berthelot, Carlini, Goodfellow, Oliver, Papernot, and Raffel]{mixUp}
David Berthelot, Nicholas Carlini, Ian Goodfellow, Avital Oliver, Nicolas Papernot, and Colin Raffel.
\newblock Mixmatch: A holistic approach to semi-supervised learning.
\newblock In \emph{Proceedings of the 33rd International Conference on Neural Information Processing Systems (NeurIPS)}. Curran Associates Inc., 2019.

\bibitem[Ho et~al.(2020)Ho, Jain, and Abbeel]{ddpm}
Jonathan Ho, Ajay Jain, and Pieter Abbeel.
\newblock Denoising diffusion probabilistic models.
\newblock In \emph{Proceedings of the 34th International Conference on Neural Information Processing Systems (NeurIPS)}, Vancouver, BC, Canada, 2020. Curran Associates Inc.

\bibitem[Islam et~al.(2024)Islam, Zaheer, Mahmood, and Nandakumar]{diffuseMix2024}
Khawar Islam, Muhammad~Zaigham Zaheer, Arif Mahmood, and Karthik Nandakumar.
\newblock Diffusemix: Label-preserving data augmentation with diffusion models.
\newblock In \emph{Proceedings of the IEEE/CVF Conference on Computer Vision and Pattern Recognition (CVPR)}, 2024.

\bibitem[Trabucco et~al.(2024)Trabucco, Doherty, Gurinas, and Salakhutdinov]{trabucco2024effective}
Brandon Trabucco, Kyle Doherty, Max~A Gurinas, and Ruslan Salakhutdinov.
\newblock Effective data augmentation with diffusion models.
\newblock In \emph{The Twelfth International Conference on Learning Representations (ICLR)}, 2024.
\newblock URL \url{https://openreview.net/forum?id=ZWzUA9zeAg}.

\end{thebibliography}
\bibliographystyle{acl_natbib}

\end{document}